\documentclass{sig-alt-release2}

\begin{document}

\conferenceinfo{GECCO'11,} {July 12--16, 2011, Dublin, Ireland.} 
\CopyrightYear{2011} 
\crdata{978-1-4503-0690-4/11/07} 
\clubpenalty=10000 
\widowpenalty = 10000

\title{Fuzzy Dynamical Genetic Programming in XCSF}

\numberofauthors{2} 

\author{
\alignauthor
Richard J. Preen\\
       \affaddr{Department of Computer Science}\\
       \affaddr{University of the West of England}\\
       \affaddr{Bristol, BS16 1QY, UK}\\
       \email{richard.preen@live.uwe.ac.uk}
\alignauthor
Larry Bull\\
       \affaddr{Department of Computer Science}\\
       \affaddr{University of the West of England}\\
       \affaddr{Bristol, BS16 1QY, UK}\\
       \email{larry.bull@uwe.ac.uk}
}

\maketitle
\begin{abstract}
A number of representation schemes have been presented for use within Learning Classifier Systems, ranging from binary encodings to Neural Networks, and more recently Dynamical Genetic Programming (DGP). This paper presents results from an investigation into using a fuzzy DGP representation within the XCSF Learning Classifier System. In particular, asynchronous Fuzzy Logic Networks are used to represent the traditional condition-action production system rules. It is shown possible to use self-adaptive, open-ended evolution to design an ensemble of such fuzzy dynamical systems within XCSF to solve several well-known continuous-valued test problems.
\end{abstract}


\section{Introduction}
Recently, we \cite{BullPreen:2009} \cite{PreenBull:2009} investigated the use of a Dynamical Genetic Programming representation scheme (DGP) within Learning Classifier Systems (LCS). It was shown that LCS are able to evolve ensembles of Random Boolean Networks (RBN) to solve a number of discrete-valued computational tasks. Additionally, it was shown possible to exploit memory existing inherently within the DGP representation. Moreover, the networks in DGP are updated asynchronously - a potentially more realistic model of Genetic Regulatory Networks (GRN) in general.

Fuzzy set theory is a generalization of Boolean logic in which continuous variables can partially belong to sets. A fuzzy set is defined by a membership function, typically within the range $[0,1]$, that determines the degree of belonging to a value of that set.

The continuous dynamical systems known as Fuzzy Logic Networks (FLN) \cite{KokWang:2006} are a generalization of RBN where the Boolean functions are replaced with fuzzy logical functions from fuzzy set theory. In this paper, we explore the use of asynchronous FLN as a representation scheme within the XCSF \cite{Wilson:2002} Learning Classifier System and show that it is possible to extend DGP to the continuous-valued domain.

\section{Fuzzy DGP in XCSF}
The following modifications are made to the discrete DGP scheme used in \cite{PreenBull:2009} to accommodate continuous-actions via fuzzy logical functions. Here, a node's function is represented by an integer which references the appropriate operation to execute upon its received inputs (see Table~\ref{table:FuzzyLogics} for the fuzzy functions used). Further, each node's connectivity is represented as a list of $k_{max}$ integers (here $k_{max}=5$) in the range $[0, N]$, where $0$ represents no input to be received on that connection. Each integer in the connection list, along with the node function, is subjected to mutation on reproduction at the self-adapting rate $\mu$ for that rule. The output nodes provide a real numbered output in the range $[0,1]$, and no averaging is used in order to preserve crisp output, however if a given FLN has a value of less than 0.5 on the match node, regardless of the state of its outputs, the rule does not join $[M]$. After building $[M]$ in the standard way, $[A]$ is built by selecting a single classifier from $[M]$ and adding matching classifiers whose actions are within a predetermined range of that rule's proposed action (here the range is set to $\pm 0.005$). Parameters are then updated as usual in $[A]$, however, similar to XCSF, the fitness adjustment takes place in $[M]$. The GA is then executed as usual in $[A]$. Exploitation functions by selecting the single rule with the highest prediction multiplied by accuracy from $[M]$. Following \cite{Wilson:2007}, an extra prediction weight, which receives as input the classifier's action, is included. In addition, the prediction weights for offspring are reset upon reproduction to prevent inexperienced rules being chosen in exploitation. 

\begin{table}[ht]
\caption{Selectable Fuzzy Logic Functions}
\centering
\begin{tabular}{l l l }
\hline\hline
ID & Function & Logic \\
\hline
0 & Fuzzy OR (Max/Min) & $max(x,y)$ \\
1 & Fuzzy AND (CFMQVS) & $x \times y$ \\ 
2 & Fuzzy AND (Max/Min) & $min(x,y)$ \\
3 & Fuzzy OR (CFMQVS and MV) & $min(1,x+y)$  \\
4 & Fuzzy NOT & $1-x$ \\
5 & Identity & $x$ \\ [0ex]
\hline
\end{tabular}
\label{table:FuzzyLogics}
\end{table}

\section{Experimentation}
\cite{Wilson:2007} presented a form of XCSF where the action was computed directly as a linear combination of the input state and a vector of action weights, and conducted experimentation on the continuous-action Frog problem, selecting the classifier with the highest prediction for exploitation. \cite{Tran:2007} subsequently extended this by adapting the action weights to the problem through the use of an Evolution Strategy (ES) and reported greater than 99\% performance after an averaged number of 30,000 trials ($P=2000$), which was superior to the performance reported by \cite{Wilson:2007}. More recently, \cite{Ramirez-Ruiz:2008} applied a Fuzzy-LCS with continuous vector actions, where the GA only evolved the action parts of the fuzzy systems, to the continuous-action Frog problem, and achieved a lower error than Q-learning (discretized over 100 elements in $x$ and $a$) after 500,000 trials ($P=200$).

The Frog Problem \cite{Wilson:2007} is a single-step problem with a non-linear continuous-valued payoff function in a continuous one-dimensional space in the range $[0,1]$. A frog is given the learning task of jumping to catch a fly that is at a distance, $d$, from the frog, where $0\le d \le 1$. The frog receives a sensory input, $x(d)=1-d$, before jumping a chosen distance, $a$, and receiving a reward based on its new distance from the fly, as given by:
\begin{equation}
P(x,a) = \left\{ \begin{array}{rl}
 x+a &\mbox{ : $x+a\le1$} \\
 2-(x+a) &\mbox{ : $x+a\ge1$}
       \end{array} \right.
\end{equation}
 
\begin{figure}[t]
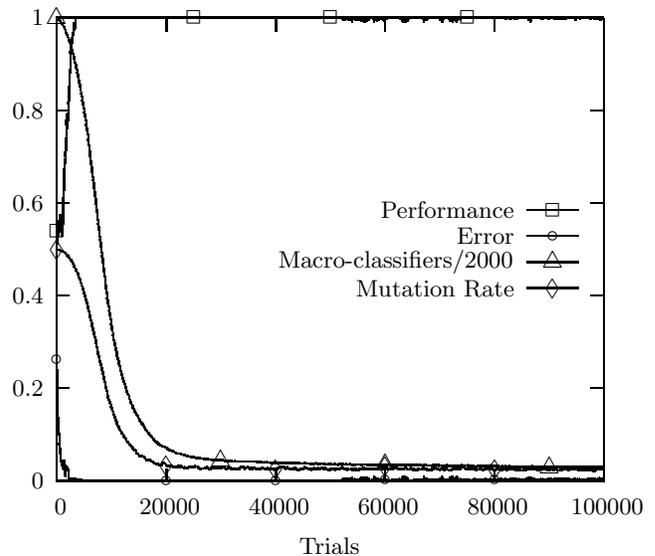

\hspace{-0.3in}
% LaTeX picture
\setlength{\unitlength}{0.240900pt}
\ifx\plotpoint\undefined\newsavebox{\plotpoint}\fi
\sbox{\plotpoint}{\rule[-0.200pt]{0.400pt}{0.400pt}}%


\caption{Performance, error, macro-classifiers and mutation rate in continuous-action Frog Problem.}
\label{fig:performance}
\end{figure}

The parameters used here are the same as used by \cite{Wilson:2007} and \cite{Tran:2007}. Fig.~\ref{fig:performance} illustrates the performance of fDGP-XCSF in the continuous-action Frog Problem. It can be seen that greater than 99\% performance is achieved in fewer than 4,000 trials ($P=2000$), which is faster than \cite{Tran:2007} (>99\% after 30,000 trials, $P=2000$) and \cite{Wilson:2007} (>95\% after 10,000 trials, $P=2000$), and with minimal changes resulting in none of the drawbacks; i.e., exploration is here conducted with roulette wheel on prediction instead of deterministically selecting the highest predicting rule, enabling true reinforcement learning. Furthermore, in \cite{Tran:2007} the action weights update component includes the evaluation of the offspring on the last input/payoff before being discarded if the mutant offspring is not more accurate than the parent; therefore additional evaluations are performed which are not reflected in the number of trials reported.

The average number of (non-unique) macro-classifiers used by fDGP-XCSF (Fig.~\ref{fig:performance}) rapidly increases to approximately 1400 after 3,000 trials, before converging to around 150; this is more compact than XCSF with interval conditions ($\sim$1400) \cite{Wilson:2007}, showing that fDGP-XCSF can provide strong generalisation. The networks grow, on average, from 3 nodes to 3.5, and the average connectivity remains static around 2.1, while the average value of $T$ increases by from 28.5 to 31.5 (not shown). The average mutation rate declines from 50\% to 2\% over the first 15,000 trials before converging to around 1.2\% (Fig.~\ref{fig:performance}).

\section{Conclusions}
It has been shown that XCSF is able to design ensembles of dynamical fuzzy logic networks whose emergent behaviour is able to be collectively exploited to solve a continuous-valued task via reinforcement learning, where performance in the continuous Frog Problem was superior to those reported previously in \cite{Ramirez-Ruiz:2008}, \cite{Tran:2007} and \cite{Wilson:2007}.

\bibliographystyle{abbrv}
%\bibliography{references} 

\end{document}